\documentclass{article}

\usepackage[final]{nips_2016}
\usepackage{subcaption}
\usepackage[utf8]{inputenc} 
\usepackage[T1]{fontenc}    
\usepackage{hyperref}       
\usepackage{url}            
\usepackage{booktabs}       
\usepackage{amsfonts}       
\usepackage{nicefrac}       
\usepackage{microtype}      
\usepackage{graphicx}
\usepackage{float}
\usepackage{mdframed}
\title{Subjective Multi-Bias Detection with Large Language Models}

\author{
  Ruiyu Li\textsuperscript{1} \\
  \texttt{ruiyul@cs.cmu.edu}
  \And
  Zhiying Zhu\textsuperscript{1} \\
  \texttt{zhiyingz@cs.cmu.edu}
}

\begin{document}

\maketitle

\section{Introduction}

In this project, we delved into the pervasive challenge of bias detection within the text content. More specifically, our focus lies on the identification of subjective bias, a type of bias that introduces improper attitudes or portrays a statement at odds with the actual truth. The subjective bias can jeopardize the authenticity and reliability of texts, leading to misconceptions and potential social tensions, especially when expressed through offensive language. 

Following prior work \cite{recasens-etal-2013-linguistic}, we tackled with three different types of subjective biases in text: (1) framing bias
with the use of one-sided words or phrases containing a particular point of view; (2) epistemological bias which includes subtle linguistic features that can affect the believability of the texts; (3) demographic bias with word/phrase usage under
presuppositions of a particular demographic factor
(i.e., gender or religion).

In terms of the data we utilize, the input consists of texts that may harbor subjective biases. The output is a classification or annotation that reveals the presence or absence of such biases within the provided content. More specifically, we detected three different types of multi-span biases in corpus WIKIBIAS \cite{zhong2021wikibias} with more than 4,000 sentence pairs from Wikipedia edits. The data is labelled by bias type for span pairs with the following categories: (1) framing bias, (2) epistemological bias, (3) demographic bias, and (4) no bias.

Going beyond the baseline BERT model, we studied a variety of cutting-edge deep learning models and more
sophisticated large language models. We decided to build and fine-tune LLaMA2-MLP classifers on our dataset to leverage its stronger language understanding ability. Moreover, we developed an ensemble approach to achieve better generalization and robustness compared to individual models. Our final model outperforms the baseline model by 26\% of macro-F1, and achieved new state-of-art results in all the classes of bias. The project codes are released at \url{https://github.com/HoningJade/LLM-Bias-Type-Classification}.

\section{Background}

Our midway work consists of implementing a BERT-based \cite{devlin-etal-2019-bert} classifer as baseline. We chose the BERT-base variant that utilizes a transformer architecture with 12 layers, 768 hidden dimensions, and 12 self-attention heads, amounting to 110 million parameters in total. We utilized pre-trained weights from the HuggingFace Transformers library \cite{wolf-etal-2020-transformers}. To detect subjective bias, we finetuned it for three epochs on the finegrained WIKIBIAS training set. The training and validation loss indicated our finetuning process is reasonable.

On the finegrained WIKIBIAS test set, the baseline achieves a macro-F1 of 0.33, which reveals the challenges of our tasks. First, some error cases require comprehensive text understanding. It's even hard for human beings to identify whether the sentences contain bias and which type of bias. To overcome BERT's limited modeling and text understanding ability, we decided to build stronger deep learning models such as LLMs \cite{radford2019language}. Secondly, the dataset is highly imbalanced. The most number of incorrect classification come from minor types, epistemological bias and demographic bias.

\section{Related Work}

\subsection{Subjective Bias Detection}
Research into the detection of subjectivity has its roots in the 1990s, initiated by the work of Karlgren and Cutting \cite{karlgren1994recognizing} and Kessler et al. \cite{kessler1997childhood}, focusing on document-level bias. This trajectory then broadened to sentence-level bias by later scholars such as Bruce and Wiebe \cite{bruce1999recognizing} and Hatzivassiloglou and Wiebe \cite{hatzivassiloglou2000effects}. Further refinement in methodology was seen with a shift towards linguistic feature-based classification(Riloff and Wiebe, \cite{riloff2003learning}, Pang and Lee, \cite{pang2004sentimental}, Lin et al., \cite{lin2011sentence}, Murray and Carenini, \cite{murray2009predicting}, Yang et al., \cite{hu2017toward}). The evolution of neural models from the mid-2000s to the 2010s, with contributions from figures like Morstatter et al. \cite{morstatter2018search}, set new standards in the field. 

\subsection{Large Language Models}

Models such as OpenAI's GPT series \cite{radford2019language} and Google's BERT \cite{devlin2018bert} have achieved groundbreaking performance across a wide range of natural language processing (NLP) tasks. Bender et al. highlighted potential societal concerns arising from these models, including their inherent biases and ethical implications \cite{bender2021dangers}. Moreover, Strubell et al. addressed their energy consumption and environmental impact \cite{strubell2019energy}. However, Chollet argued that LLMs, while advanced, still lack true understanding and reasoning abilities \cite{chollet2019measure}.

With the advent of large language models, our motivation is geared towards leveraging their capabilities to detect subjective bias. This not only enhances the granularity of detection but also integrates the comprehensive knowledge of these models.  

\subsection{Ensemble learning}
The field of ensemble learning has witnessed significant attention and advancement in recent years, owing to its ability to enhance predictive performance and robustness of machine learning models \cite{9893798}. Noteworthy contributions include research on diverse ensemble methods such as bagging, boosting, and stacking, each offering unique advantages in handling different aspects of model uncertainty and bias \cite{9170675}\cite{ganaie2022ensemble}. 

Some recent work introduced ensemble learning algorithms to solve the class imbalance problem in deep learning, such as \cite{ahmed2017hybrid} and \cite{Malek2022ComparisonOE}. \cite{chen2021class} proposes an ensemble of auxiliary classifiers branching out from various hidden layers of a CNN. They designed a new loss function that can rectify the bias toward the majority classes by forcing the CNN's hidden layers and its associated auxiliary classifiers to focus on the samples that have been misclassified by previous layers, thus enabling subsequent layers to develop diverse behavior. \cite{Malek2022ComparisonOE} compared the performance of random forest, gradient boosting, and three sampling approaches to solve class imbalance on water quality data. According to their results, the best model was gradient boosting without resampling.

\section{Methods}

\subsection{Data and Preprocessing}
The WikiBIAS dataset \cite{zhong2021wikibias} is an innovative, high-quality parallel corpus meticulously designed to delve into subjective biases in text. Originating from Wikipedia edits, this dataset consists of over 4,000 manually annotated sentence pairs, covering both sentence-level biases and detailed, token-level biases. Rooted in over 53.5k non-identical word alignments, WikiBIAS surpasses the boundaries of prior works by employing a meticulous two-stage annotation methodology, ensuring unparalleled accuracy in bias detection.

About preprocessing, the dataset is split into 5,028 training samples, 1,066 validation samples, and 2,104 test samples. We chose the WikiBIAS dataset for our project due to its unique focus on multi-word and multi-span subjective bias. Seamlessly integrating automatic labeling with human expertise, WikiBIAS emerges as a cornerstone tool for in-depth bias analysis in textual data.

\subsection{Baseline Model}

We implement the BERT-base model \cite{devlin2018bert} as a baseline model. Specifically, the BERT-base variant utilizes a transformer architecture with 12 layers, 768 hidden dimensions, and 12 self-attention heads, amounting to 110 million parameters in total. To pretrain, BERT uses a combination of masked language modeling and next sentence prediction tasks on large corpora like Wikipedia and BooksCorpus. For detecting subjective bias, we fine-tuned the pretrained BERT-base model on WikiBIAS. The first token of every sequence is a special [CLS] token, and we used this as the sentence representation for classification. The output layer is a simple linear layer that outputs class probabilities, and the model is trained to minimize the cross-entropy loss between the predicted and true labels.
    
\subsection{Learning EnsembleLlama Experts}
WIKIBIAS is a fairly new and challenging benchmark. Despite the significance of subjective bias with its wide applications to news, political speeches, and many other domians, existing works struggle with detecting multi-span biases in it. To this end, we came up with our original idea of developing and LLama2-based deep classifiers and stacking the learned experts to mitigate data imbalance.
\subsubsection{Learning LLama2-MLP classifiers}
To develop models with stronger capability of language sequence understanding, we researched into and compared various LLMs. We found that successful scaling of LLMs has substantially improved model performance and exhibited emergent abilities in addressing complex NLP tasks. Among existing LLMs, GPT-3.5~\cite{NEURIPS2020_1457c0d6} and LLaMA2-70B~\cite{Hugo2023} are decoder-only models and leverage a Transformer decoder architecture. The models have vast number of parameters, making fine-tuning for specific tasks very challenging. An alternative to finetuning is to craft few-shot in-context examples and provide them as input context during inference. \cite{brown2020language} showed that LLMs are capable of learning from these examples to improve over zero-shot performance.

\begin{figure}[h]
    \centering
    \includegraphics[width=0.3\textwidth]{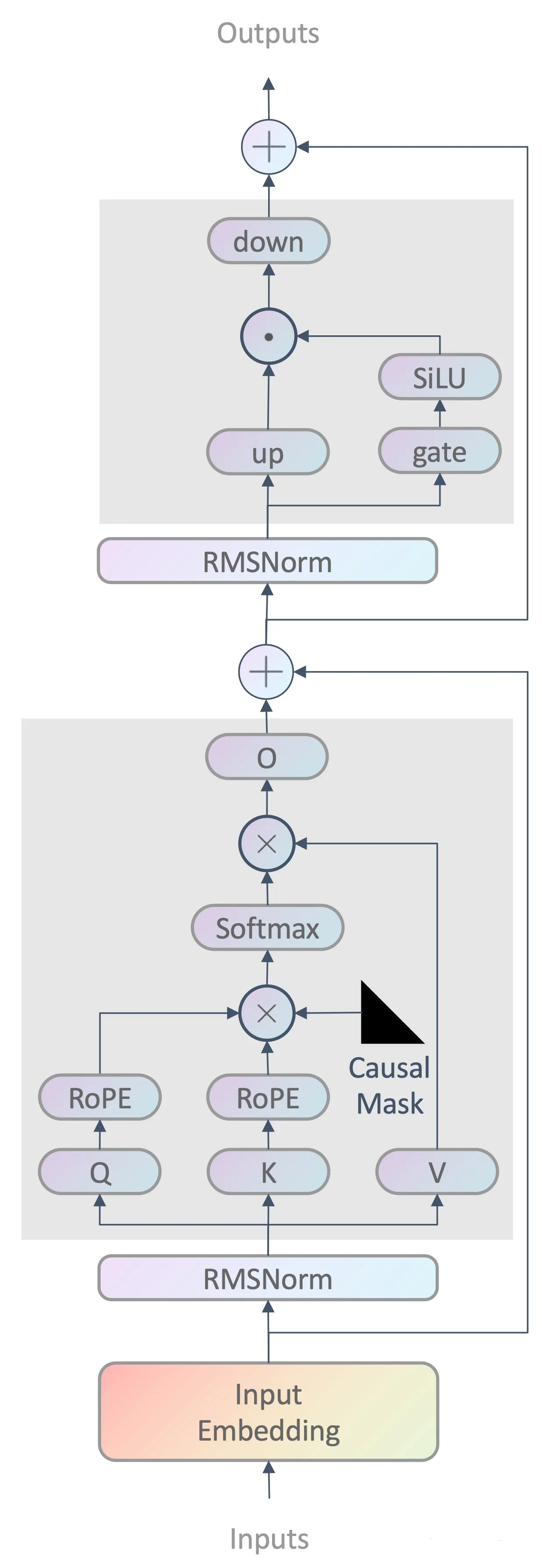}
    \caption{LLama model architecture \cite{zhihu}}
    \label{llama}
\end{figure}

We decided to develop a classifier upon LLama 2's architecture\cite{Hugo2023} because (1) it achieved cutting-edge performance on most benchmarks, (2) it's open-source, (3) its released 7B pretrained model is of practical size for us to finetune, and (4) we are able to further develop a classifer based on it to train for our tasks. The LLaMA 2 model are largely based on the Transformers architecture. In addition, the following improvements are added to LLaMA 2. First, LLaMA adopted pre-normalization, which gives the models improvements for re-scaling invariance property and implicit learning rate adaptation ability. Secondly, in LLaMA, SwiGLU activation function replaced the ReLU non-linearity and achieved significantly improvement over transformer models. Moreover, the absolute positional embeddings are removed and instead rotary positional embeddings are added. Figure \ref{llama} shows an architecture of LLaMA-2. Table \ref{llama2} shows key settings and details of LLama2-7B.
\begin{table}[h]
    \centering
    \begin{tabular}{lccccc}
        \hline
        Training Data & Params & Context Length & GQA & Tokens & LR \\ \hline
        Mix of publicly available online data & 7B   & 4k & No & 2T & 3$\times 10^{-4} $\\
        \hline
    \end{tabular}
    \caption{Key settings and details of Llama2-7b}
    \label{llama2}
\end{table}

To enable fine-tune LLama-2 for our tasks, we added linear layers at the end of LLaMA-2 and trained the linear layers on bias type classification dataset. We freeze LLama-2 with  pre-trained weights, so our approach is computation-efficient and don't cost huge amount of resources. Besides, we designed three system messages as an initial prompt to interact with LLama-2. In these prompts , we first explain to LLMs the concept of certain bias type and then give an example to better illustrate the concept. Below is our designed prompt for LLama-2 to detect framing bias.
\begin{mdframed}
Framing bias is a type of subjective bias that uses of one-sided words or phrases containing a particular point of view. Here is an example: New York is the greatest state in the northeastern United States . Does the following sentence have framing bias? 
\end{mdframed}
Below is our designed prompt for LLama-2 to detect epistemological bias.
\begin{mdframed}
Epistemological bias is a type of subjective bias which includes subtle linguistic features that can affect the believability of the texts. Here is an example: People with disabilities are excluded from cultural and social norms that afford pleasure ( sex , sex education , sexual health , marriage ) . Does the following sentence have epistemological bias?  
\end{mdframed}
Below is our designed prompt for LLama-2 to detect demographic bias.
\begin{mdframed}
Demographic bias is a type of subjective bias with word/phrase usage under presuppositions of a particular demographic factor (i.e., gender or religion). Here is an example: Complementarian an alternative Christian view that interprets scripture to teach that women and men as created equal though men are to hold ultimate authority over women in Church and the home . Does the following sentence have demographic bias? 
\end{mdframed}
\subsubsection{Stacking experts to address class imbalance}

We first trained the LLama-MLP to output three logits for each input and each logit correspond whether one of the three types of biases is present in the input, which follows the way of the benchmark work. We also trained the LLama-MLP to directly output a label between 0 to 7 inclusively for each sentence since the combination of possible categories of each input is eight. However, both of the experiments give similar performance as the baseline because they encountered the class imbalance problem mentioned before. To address this, we tried asymmetric loss and focal loss \cite{lin2017focal} to improve performance, but we found that the class imbalance still presents and the model misclassified many minor classes. 

We conducted human evaluation where we looked at input texts and identified biases by ourselves. We don't see much correlation among the three types of biases themselves. Thus, we further creatively re-formulate the  output and fine-tuned three LLama-MLP classifiers, each of which only focus on detecting one of the biases. This allows us to effectively address class imbalance for each classifier. We pre-processed the dataset to separate labels for different biases and to make the dataset compatible for our new approach. We stack the detection results from the three LLama-MLP experts to give the final detection results on three bias types. We studied various ensemble algorithms for deep learning and in our experiments just stacking can boost performance greatly. In this way, our final outputs are of the same formulation as the benchmarks.

\section{Results}
We implemented our EnsembleLlama via the HuggingFace meta-llama/Llama-2-7b-hf repository \cite{huggingface} and FastChat\cite{zheng2023judging} , an open platform for training, serving, and evaluating large language model based chatbots. We finetuned it for three epochs on the finegrained WIKIBIAS training set. We used the finegrained WIKIBIAS validation set for our baseline validation and tested the baseline on the finegrained WIKIBIAS testing set. For the hyperparameters in finetuning, we set batch size as 2 and max length as 2048. For learning rate, we set initial learning rate as 2e-5 and used a cosine annealing learning rate scheduler. We ran experiments on two NVIDIA 80GB H100 GPU and three epochs of training takes about 1 hour. 

We report the macro and class-level F1 of our EnsembleLLama and the baseline on test set in Table \ref{tab:experiment_results}. Our EnsembleLLama achieves a macro-F1 of 0.59, which is 0.26 higher than the baseline. We also significantly improved class-level F1 of all the classes by about 0.25. To sum up, our new model obtains state-of-art results on WIKIBIAS benchmark.

\begin{table}[h]
    \centering
    \begin{tabular}{lcccc}
        \hline
        Model & macro-F1 & F-class F1 & E-class F1 &  D-class F1 \\ \hline
        Finetuned BERT      & 0.33   & 0.49 &0.25&    0.25   \\ \hline
        EnsembleLlama (ours)      & \textbf{0.59}   & 0.75 &0.52&    0.49  \\\hline
        Our improvement    & \textbf{0.26}  & 0.26 & 0.27 &    0.24  \\
        \hline
    \end{tabular}
    \caption{Result Comparison with Macro and class-level F1}
    \label{tab:experiment_results}
\end{table}

\begin{figure}[hbt!]
     \centering
     \begin{subfigure}[b]{0.48\textwidth}
         \centering
         \includegraphics[width=1.1\textwidth]{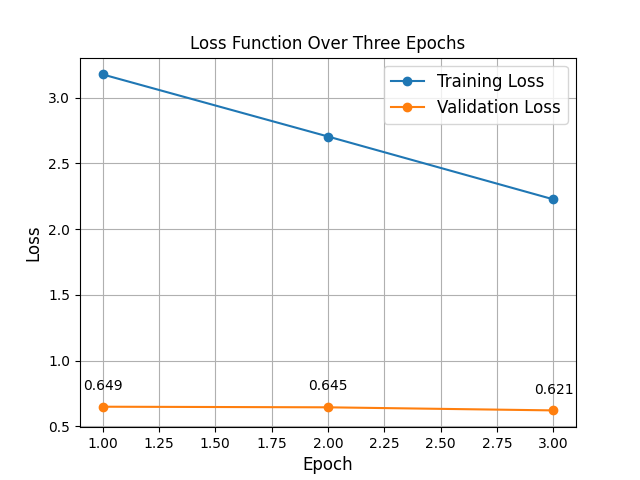}
         \caption{Loss}
         \label{loss}
     \end{subfigure}
     \hfill
     \begin{subfigure}[b]{0.48\textwidth}
         \centering
    \includegraphics[width=1.1\textwidth]{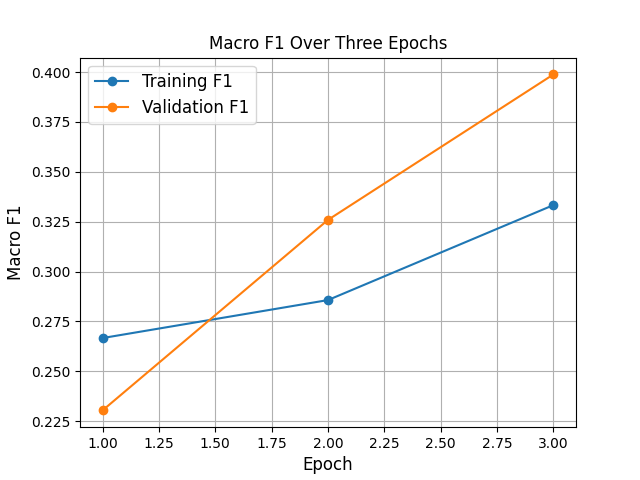}
         \caption{Macro F1}
         \label{f1}
     \end{subfigure}
        \caption{Training and validation results of the fintuned baseline}
        \label{fig:three graphs}
\end{figure}

\begin{figure}[hbt!]
     \centering
     \begin{subfigure}[b]{0.65\textwidth}
         \centering
         \includegraphics[width=1.1\textwidth]{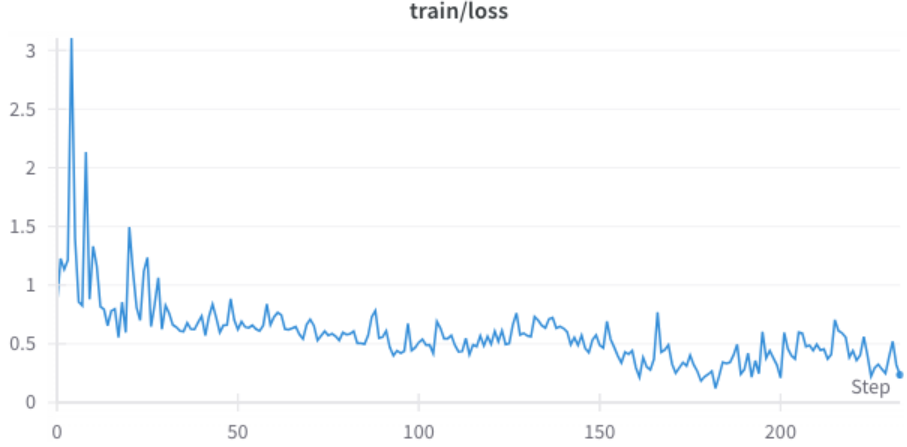}
         \caption{F-class Loss}
         \label{loss}
     \end{subfigure}
     \hfill \\
     \begin{subfigure}[b]{0.65\textwidth}
         \centering
    \includegraphics[width=1.1\textwidth]{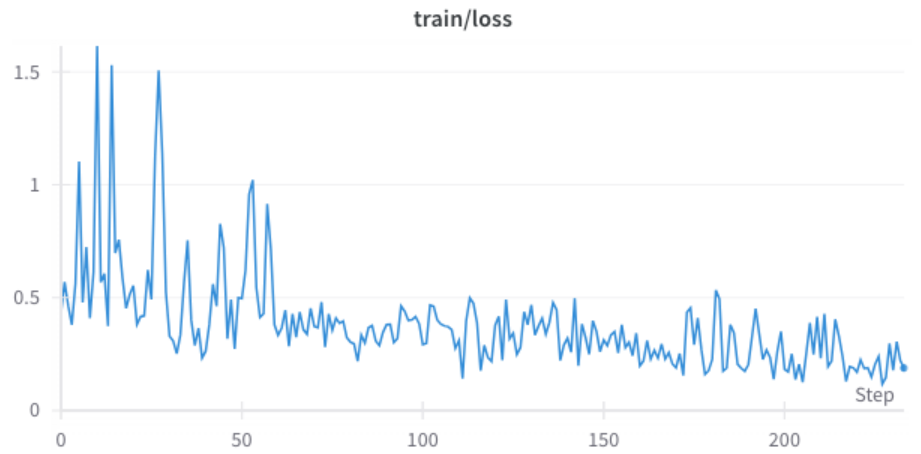}
         \caption{E-class Loss}
         \label{f1}
     \end{subfigure}
     \hfill \\
     \begin{subfigure}[b]{0.65\textwidth}
         \centering
    \includegraphics[width=1.1\textwidth]{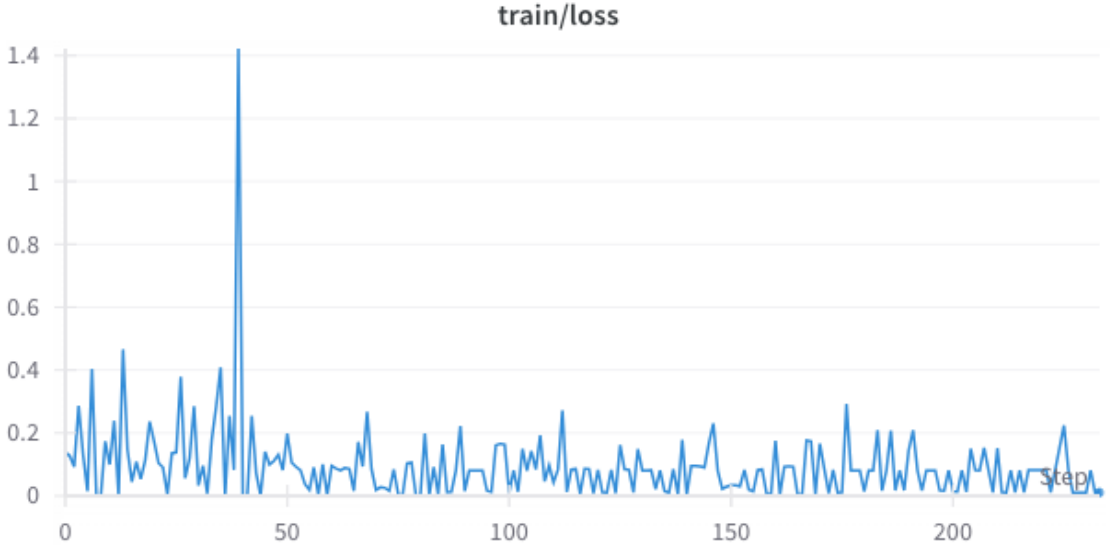}
         \caption{D-class Loss}
         \label{f1}
     \end{subfigure}
        \caption{Loss curves of training LLaMA-MLP classifiers}
        \label{finalfig}
\end{figure}

We also visualized the loss and macro-F1 on the training set and validation set during fine-tuning the models in Fig.\ref{fig:three graphs} and Fig. \ref{finalfig}. We can see that the loss gradually decreases and the macro F1 increases as the number of epochs/steps increases. There is no overfitting. 

Examining model outputs on test set, we found that our model effectively corrected many previous error cases. In the box below, we show an example in test set that is misclassied by the baseline model but our EnsembleLLama performed correct detection.
\begin{mdframed}
\centering Test Set Example \\
Input text: Three people were killed and more than 60 wounded , one of the dead being IRA volunteer Caoimhn Mac Brdaigh .
\\Label: epistemological bias
\\Baseline result: framing bias
\\Our EnsembleLlama result: epistemological bias
\end{mdframed}

\section{Discussion and Analysis}

    \subsection{Baseline models}
While the BERT-based model utilizes a multi-layer bidirectional Transformer to obtain deep bidirectional representations from  text, its performance on our task is low. This is due to (1) imbalance data and many incorrect classification of epistemological and demographic bias; (2) BERT's limited modeling ability. To solve the problems, we plan to (1) design a loss to assign different weights to different classes (2) build stronger deep learning models such as RoBERTa\cite{liu2019roberta} and LLMs \cite{radford2019language}. 

We analyzed the model outputs and identified common errors. (1) The model misclassified sentences as framing bias. For example, the epistemological bias in "Three people were killed and more than 60 wounded , one of the dead being IRA volunteer Caoimhn Mac Brdaigh.", is classified as framing bias. (2) When a sentence has multiple biases, the model missed some of the bias. For example, "Tofurkey is ideal as the main dish of a formal meal for people who have ethical reasons to abstain from eating meat.",  has both farming and demographic biases, but the model prediction missed demographic bias. 
    
    \subsection{EnsembleLlama Classifiers}
While our approach boosted the detection performance on each of the class, the F-class F1 is 1.4 times as high as the E-class F1 and D-class F1. As a comparison, the class-level F1 of F-class is about twice as high as E-class and D-class in the baseline. Hence, our approach have mitigated class imbalance by removing the influence of dominant bias on minor biases. But still, the imbalance between minor biases and no bias impedes model performance.

While we achieved twice as high macro-F1 as the baseline, our model performance is still limited. Below is an error case of our model on the test set. We can see that it requires deep understanding of the text and solid knowledge to judge whether the text contains framing bias or not. So another limitation of our model is that its performance highly depends on the text understanding ability and the knowledge of LLama2. We could potentially equip strong LLM with domain knowledge on subjective bias to create specialized expert models with better performance in the future.

\begin{mdframed}
Bancroft was founded in 1900 by a group of Worcester parents interested in providing a rigorous education for their children . \\
Label: framing bias\\
Our result: no bias
\end{mdframed}

In addition, we recognize  that simply stacking individual classifiers to obtain final detection in our model may not fully utilize the advantages and possible mutual promotion of separate classifiers. 

Our present methodology is centered around the identification of sentence-level bias. While this approach yielded significant insights, we recognize the potential for more granular analyses. We believe that bias often manifests in subtle ways, potentially at the level of individual word choices or edits. 

    \subsection{Future Work}
Based on the above analysis, we identified four primary areas of future work that will enhance the depth and breadth of our study.
    
    We plan to explore more advanced methods to further address class imbalanced in deep learning. For example, we will implement sampling, data
    augmentation, staged learning, and model design methods discussed in \cite{henning-etal-2023-survey} to further improve our task performance.

    We will try more advanced ensemble learning methods to further improve model performance. We plan to try the ensemble learning algorithms for deep learning mentioned in related work to better aggregate the results from individual classifiers.
    
    Moreover, we plan to evolve our algorithms and methods to detect and understand biased language at this intricate single-word edit level. This will not only enhance the precision of our bias detection but also provide a nuanced understanding of how biases are embedded in textual content.

    We also want to extend our method on more datasets and benchmarks of subjective bias detection to evaluate its generalization ability and for more extensive experiments.

\newpage
\bibliographystyle{unsrt}
\bibliography{refs}

\end{document}